\documentclass[11pt]{article}
\usepackage[margin=1in]{geometry}
\usepackage{times}

\usepackage{amsmath,amssymb}
\usepackage{booktabs}
\usepackage{graphicx}
\usepackage{microtype}
\usepackage{multirow}
\usepackage{xcolor}
\usepackage{hyperref}
\usepackage{url}
\usepackage[numbers,sort&compress]{natbib}
\hypersetup{
  colorlinks=true,
  linkcolor=black,
  citecolor=black,
  urlcolor=blue
}

\newcommand{\method}{\textsc{DSF}}
\newcommand{\R}{\mathbb{R}}
\newcounter{algorithm}
\renewcommand{\thealgorithm}{\arabic{algorithm}}

\title{Dynamic Spectral Filtering for Temporal Graph Learning:\\
Learning Evolving Propagation Operators}

\author{
Yan Kong\\
Nanjing University of Information Science and Technology, China\\
\texttt{kongyan4282@163.com}
}
\date{}

\begin{document}
\maketitle

\begin{abstract}
Temporal graph learning is commonly organized around the evolution of node
states or the encoding of interaction histories. We study an underexplored,
operator-centric question: should the graph propagation mechanism itself
evolve over time?
We introduce Dynamic Spectral Filtering (\method), which represents propagation
at snapshot $t$ by a Chebyshev polynomial filter with vector-valued,
time-dependent coefficients. \method{} explicitly treats these compact
multi-order coefficients as recurrent temporal states. A recurrent branch proposes
updates, while multiplicative global and order-specific gates regulate their
magnitude. The temporal state is independent of the number of nodes. On MOOC,
Wikipedia, and Reddit temporal link-prediction benchmarks, converged \method{}
runs attain AP scores of 0.7851, 0.9088, and 0.9860, respectively, with
93K--133K trainable parameters, 68--182 MB peak GPU memory, and 1.6--2.1
seconds of training per epoch. Against the closely related DEFT baseline,
\method{} is better on MOOC, within 0.001 AP on Reddit, and modestly lower on
Wikipedia, while using $8.3$--$8.6\times$ fewer parameters, $25$--$33\times$
less GPU memory, and $5$--$19\times$ less time per epoch. Relative to all
measured alternatives, it uses $3.3$--$38.6\times$ less GPU memory. These
results support direct spectral-response evolution as a useful temporal
inductive bias when computational efficiency is a first-class requirement.
\end{abstract}

\section{Introduction}

Graphs observed in recommender systems, online communities, communication
networks, and transaction platforms are not static objects. Edges arrive over
time, neighborhoods change, and the relevance of local versus multi-hop
information can vary across stages of the process. Temporal graph learning
therefore requires an inductive bias not only for relational structure, but
also for how that structure should be used at each time.

Two prominent families place temporal variation in node-associated states or
interaction histories.
Memory-based event models such as JODIE and TGN update node-associated states
as interactions arrive \cite{kumar2019jodie,rossi2020tgn}. Attention-based
models such as TGAT and DyGFormer instead retrieve and encode temporally ordered
interaction neighborhoods \cite{xu2020tgat,yu2023dygformer}. Although these
families differ substantially, both ultimately produce time-dependent node
representations. Other work evolves GNN weight matrices or spectral
representations directly \cite{pareja2020evolvegcn,bastos2023deft,bastos2024eft},
showing that temporal adaptation can also occur inside the graph computation.

\begin{figure}[t]
  \centering
  \includegraphics[width=\linewidth]{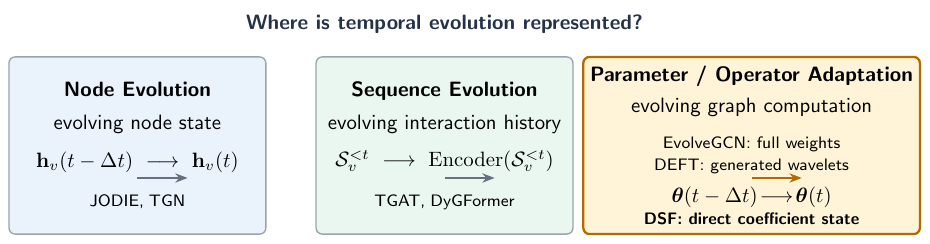}
  \caption{Illustrative locations for temporal evolution in dynamic graph
  learning; these are modeling perspectives rather than an exhaustive or
  historical taxonomy. EvolveGCN adapts full GCN weights and DEFT generates
  time-specific wavelet filters inside a composite spectral--spatial model.
  \method{} directly treats compact multi-order propagation coefficients as
  the temporal state.}
  \label{fig:paradigm}
\end{figure}

We investigate a more specific, operator-centric design axis:
\emph{compact spectral-response evolution}. Our central hypothesis is that
temporal change can alter not only node states and observed histories, but also
the appropriate rule for propagating information.
For example, a recent-neighbor signal may dominate during a burst of activity,
whereas a smoother, broader neighborhood signal may be more useful during a
stable period. A fixed spectral response cannot express this change directly.
Figure~\ref{fig:paradigm} places this perspective alongside node, sequence,
and earlier parameter/operator adaptation approaches.

To instantiate this idea, we propose Dynamic Spectral Filtering (\method{}).
At time $t$, \method{} constructs a polynomial graph filter
\begin{equation}
  \mathcal{F}_t(X'_t)
  =
  \sum_{k=0}^{K}
  \left[T_k(L_t)X'_t\right]\odot\boldsymbol\theta_{k,t},
  \label{eq:intro_operator}
\end{equation}
where $T_k$ is the $k$th Chebyshev polynomial of the normalized graph
Laplacian and $\boldsymbol\theta_{k,t}\in\R^d$ is a time-dependent,
channel-wise response for order $k$. A compact recurrent state generates
proposed coefficient updates, and a hierarchical gate regulates global and
order-level change. Thus, temporal modeling occurs directly in the propagation
rule.

This placement is also computationally consequential. Rather than maintaining
a temporal state for every node or repeatedly attending over long interaction
histories, \method{} carries $(K+1)d$ evolving spectral coefficients plus a
fixed-dimensional recurrent state. In our setting, $K=3$ and $d=64$, so the
propagation response contains only 256 values, independent of graph size. As shown in
Table~\ref{tab:resource}, this design translates into consistently lower
parameter counts, peak GPU memory, and measured training time than the
evaluated TGN, TGAT, DyGFormer, and DEFT implementations. In particular,
\method{} remains close to the substantially heavier DEFT spectral baseline in
predictive performance while occupying a markedly better
accuracy--efficiency region.

Our contributions are:
\begin{itemize}
  \item We formulate operator evolution as a focused temporal modeling
  principle: temporal adaptation can act directly on the rule governing
  information propagation, rather than exclusively on node representations or
  histories.
  \item We introduce \method{}, which explicitly models compact multi-order
  spectral coefficients as recurrent temporal states and directly evolves them
  through global and order-specific hierarchical gates.
  \item We show that the temporal operator state is compact and independent of
  graph size, yielding 93K--133K parameters, 68--182 MB peak GPU memory, and
  1.6--2.1 seconds per training epoch in our three benchmark runs---up to
  $27.6\times$ fewer parameters and $38.6\times$ less peak GPU memory than the
  measured alternatives.
  \item Against the closely related official DEFT implementation, \method{}
  achieves comparable predictive performance while using approximately
  $8.3$--$8.6\times$ fewer parameters, $25$--$33\times$ less GPU memory, and
  $5$--$19\times$ less training time per epoch.
\end{itemize}

\section{Related Work}

\paragraph{Temporal node-state models.}
JODIE co-evolves user and item embeddings after interactions
\cite{kumar2019jodie}. TGN provides a general event-based framework combining
node memories, messages, and graph operators \cite{rossi2020tgn}. These methods
make node-associated representations the principal carrier of temporal state.

\paragraph{Temporal sequence and neighborhood encoders.}
TGAT applies temporal attention with functional time encodings
\cite{xu2020tgat}. CAWN represents temporal motifs through causal anonymous
walks \cite{wang2021cawn}, while DyGFormer uses patch-based interaction
sequences to model longer histories efficiently \cite{yu2023dygformer}.
The common emphasis is on retrieving and encoding temporal interaction
contexts.

\paragraph{Parameter evolution in dynamic GNNs.}
EvolveGCN uses a recurrent network to evolve the weight matrices of a GCN
rather than maintaining node embeddings \cite{pareja2020evolvegcn}. This work
is an important precedent for placing temporal state in the model parameters.
\method{} takes a narrower spectral route: it keeps the feature transformation
shared and evolves a compact bank of Chebyshev coefficients, with separate
global and spectral-order gates controlling each residual update. The resulting
state parameterizes the propagation response directly rather than evolving the
full graph-convolution weight matrix.

\paragraph{Spectral learning on dynamic graphs.}
Spectral graph convolution defines graph propagation through functions of the
graph Laplacian. ChebNet replaces eigendecomposition with localized Chebyshev
polynomials, enabling linear-time filtering in the number of edges
\cite{defferrard2016chebnet}. DEFT learns time-specific spectral graph wavelet
coefficients through an evolving GNN, pooling, and an MLP, and combines the
resulting spectral branch with spatial message passing and representation
integration \cite{bastos2023deft}. The Evolving Fourier Transform instead
learns an invertible time-aware spectral basis \cite{bastos2024eft}. These
works rule out any claim that \method{} is the first dynamic spectral model.
The distinction is narrower and architectural: DEFT indirectly generates
wavelet filters through an evolving feature encoder, whereas \method{}
directly maintains the compact multi-order response as a recurrent temporal
state and makes this operator evolution the primary temporal mechanism. Thus
our claim is not ``spectral versus non-spectral'' or historical priority, but
a focused, explicit, and resource-efficient realization of coefficient-state
evolution.

\section{Dynamic Spectral Filtering}

\begin{figure}[t]
  \centering
  \includegraphics[width=\linewidth]{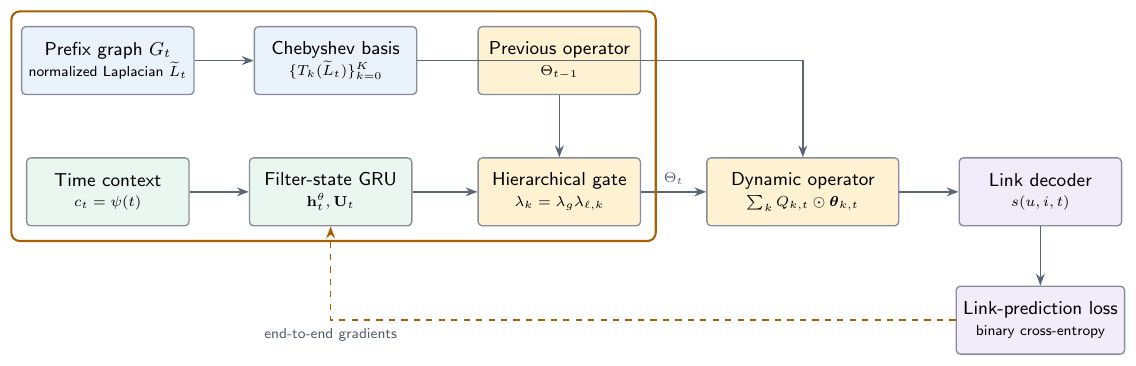}
  \caption{\method{} at snapshot $t$. The observed graph prefix determines the
  Chebyshev basis. A recurrent temporal branch proposes vector-valued filter
  updates, and multiplicative global and order-specific gates determine the
  effective update magnitude. Link-prediction loss trains the recurrent,
  gating, propagation, embedding, and decoder modules jointly.}
  \label{fig:architecture}
\end{figure}

\subsection{Temporal Link Prediction}

Let $\{(u_i,v_i,t_i,\mathbf e_i)\}_{i=1}^{M}$ be a time-ordered interaction
stream, where $\mathbf e_i\in\R^{d_e}$ is an observed interaction feature
provided by the dataset rather than a learned model embedding. We partition
the stream into $T$ nonempty temporal buckets. The prefix
graph available immediately before bucket $t$ is
$G_t=(V,A_t,X_t)$. Its node features aggregate historical edge features and
two node-type indicators. Current-bucket interactions are appended only after
their scores are computed.

More precisely, let
$\mathcal H_v(t)=\{j:t_j<t,\ v\in\{u_j,v_j\}\}$ be the visible interaction
history of node $v$. The input feature is
\begin{equation}
 \mathbf x_v(t)=
 \left[
 \overline{\mathbf e}_v(t);
 \mathbb I(v\text{ is a user});
 \mathbb I(v\text{ is an item})
 \right],
 \qquad
 \overline{\mathbf e}_v(t)=
 \frac{1}{|\mathcal H_v(t)|+1}\sum_{j\in\mathcal H_v(t)}\mathbf e_j,
 \label{eq:node_input}
\end{equation}
where the extra denominator count is the implementation's zero-vector prior
and also yields a zero vector when the visible history is empty. Hence node
identifiers are scalar indices in the event table, while
$\mathbf x_v(t)\in\R^{d_e+2}$ is the multivariate model input.

\subsection{Vector-Valued Spectral Propagation}

Let $L_t=I-D_t^{-1/2}A_tD_t^{-1/2}$ be the normalized Laplacian used by the
implementation. After an input
projection $X'_t=X_tW_x\in\R^{|V|\times d}$, the Chebyshev feature terms are
\begin{align}
  Q_{0,t}&=X'_t,\qquad Q_{1,t}=L_tX'_t, \nonumber\\
  Q_{k,t}&=2L_tQ_{k-1,t}-Q_{k-2,t}.
  \label{eq:cheb_recurrence}
\end{align}
Unlike a scalar polynomial coefficient, \method{} maintains one channel-wise
coefficient vector $\boldsymbol\theta_{k,t}\in\R^d$ for each order. Propagation
is
\begin{align}
  Z_t &= \sum_{k=0}^{K}
  Q_{k,t}\odot\boldsymbol\theta_{k,t}, \label{eq:dsf_filter}\\
  H_t &= \operatorname{ReLU}(Z_t), \label{eq:dsf_layer}
\end{align}
where each $\boldsymbol\theta_{k,t}$ is broadcast over nodes. Thus
$\boldsymbol\Theta_t\in\R^{(K+1)\times d}$ explicitly parameterizes the
time-dependent propagation operator.

\subsection{Recurrent Filter State}

The normalized snapshot index is $\tau_t=t/(T-1)$. A context encoder and GRU
produce the global filter-evolution state:
\begin{align}
  \mathbf c_t &= \operatorname{ReLU}(W_c\tau_t+\mathbf b_c),\\
  \mathbf h_t^\theta &=
  \operatorname{GRUCell}(\mathbf c_t,\mathbf h_{t-1}^\theta).
  \label{eq:filter_gru}
\end{align}
The proposed coefficient update is
\begin{equation}
  \mathbf U_t
  =
  \operatorname{reshape}\!\left(
  W_U\mathbf h_t^\theta+\mathbf b_U,(K+1,d)
  \right).
  \label{eq:proposed_update}
\end{equation}
For the first snapshot, $\boldsymbol\Theta_{t-1}$ is initialized by a small
time-conditioned MLP. The recurrent branch receives time context only; it does
not maintain node-level memory or access future interactions.

\subsection{Hierarchical Evolution Gate}

Let $\mathbf q_t=[\mathbf h_t^\theta;\mathbf c_t]$. A global gate controls
overall operator change, while a local gate assigns a relative rate to each
Chebyshev order:
\begin{align}
  \lambda_g(t)
  &=\sigma\!\left(\operatorname{MLP}_g(\mathbf q_t)\right)\in(0,1),\\
  \boldsymbol\lambda_\ell(t)
  &=\sigma\!\left(\operatorname{MLP}_\ell(\mathbf q_t)\right)
  \in(0,1)^{K+1},\\
  \boldsymbol\lambda(t)
  &=\lambda_g(t)\boldsymbol\lambda_\ell(t).
  \label{eq:hierarchical_gate}
\end{align}
The exact residual update is
\begin{equation}
  \boldsymbol\theta_{k,t}
  =
  \boldsymbol\theta_{k,t-1}
  +\alpha\,\lambda_g(t)\lambda_{\ell,k}(t)
  \mathbf U_{k,t},
  \qquad k=0,\ldots,K,
  \label{eq:gated_update}
\end{equation}
with fixed step scale $\alpha=0.1$. Multiplication makes the hierarchy
explicit: an order can change rapidly only when both the system-level and
order-level gates permit it. The gates are outputs of parameterized neural
modules, not hyperparameters. Their parameters, the recurrent update branch,
the spectral encoder, item embeddings, and decoder are optimized jointly
end-to-end by the link-prediction loss.

\subsection{Link Decoder and Complexity}

For candidate edge $(u,i)$, the decoder uses
\begin{align}
  \mathbf r_{u,i,t}
  &=[H_{u,t};H_{i,t};\mathbf p_i],\\
  s(u,i,t)&=
  W_2\operatorname{ReLU}(W_1\mathbf r_{u,i,t}+\mathbf b_1)+b_2,
  \label{eq:decoder}
\end{align}
where $\mathbf p_i\in\R^{d_p}$ is a learned candidate-item embedding, distinct
from the observed interaction feature $\mathbf e_j$. The embedding dimension
$d_p$ is a configurable model width (set to 32 in our experiments), not part
of the general formulation. Training uses binary cross-entropy with logits
over positive and sampled negative edges; its gradients jointly update all
trainable modules.

The Chebyshev recurrence requires $O(K|E_t|d)$ sparse propagation per snapshot.
The evolving operator state has $(K+1)d$ coefficients and a fixed-dimensional
GRU state, independent of $|V|$. This contrasts with node-indexed temporal
memory of size $O(|V|d)$, although full-snapshot propagation still scales with
the observed graph. With the adopted $K=3$ and $d=64$, only 256 values define
the time-varying spectral response. The model therefore spends temporal
capacity on how information propagates, rather than allocating a separate
persistent state to every node.

\subsection{End-to-End Procedure}

\refstepcounter{algorithm}
\label{alg:dsf}
\begin{center}
\fbox{\begin{minipage}{0.92\linewidth}
\textbf{Algorithm \thealgorithm: Dynamic Spectral Filtering}\\[2pt]
\textbf{Input:} ordered buckets $\{\mathcal B_t\}_{t=1}^{T}$ and order $K$.
\begin{enumerate}
  \item Initialize prefix graph, $\mathbf h_0^\theta$, and
  $\boldsymbol\Theta_0$.
  \item For each $t=1,\ldots,T$:
  \begin{enumerate}
    \item Build $L_t$ from interactions preceding $\mathcal B_t$ and compute
    $\{Q_{k,t}\}_{k=0}^{K}$.
    \item Update $\mathbf h_t^\theta$ and
    $\mathbf U_t$ using
    Equations~\eqref{eq:filter_gru}--\eqref{eq:proposed_update}.
    \item Compute the hierarchical gates and
    $\boldsymbol\Theta_t$ using
    Equations~\eqref{eq:hierarchical_gate}--\eqref{eq:gated_update}.
    \item Encode nodes, score links, and append $\mathcal B_t$ to the prefix.
  \end{enumerate}
\end{enumerate}
\textbf{Output:} temporally indexed link scores.
\end{minipage}}
\end{center}

\section{Experiments}
\label{sec:experiments}

\subsection{Research Questions}

We ask: (RQ1) Can an evolving spectral response support temporal link
prediction across datasets? (RQ2) Does the compact operator state yield a
favorable parameter and memory profile? (RQ3) How should the resulting numbers
be interpreted when snapshot and event-stream implementations use different
sampling granularities? (RQ4) How does direct coefficient-state evolution
compare with DEFT, the closest dynamic spectral baseline?

\subsection{Datasets, Metrics, and Configuration}

We use MOOC, Wikipedia, and Reddit. Each \method{} run constructs 50 nonempty
temporal snapshots, uses a 70/15/15 split by snapshot index, and makes the graph
undirected. We report test AUC and average precision (AP). All adopted
\method{} results use seed 0, a 100-epoch budget, and checkpoint selection by
validation AP only; test AP is never used for selection.

Across datasets, the hidden, temporal-context, GRU, and gate dimensions are 64;
the Chebyshev order is $K=3$; the update scale is $\alpha=0.1$; and dropout is
zero. We optimize with AdamW using learning rate $10^{-3}$, weight decay
$10^{-4}$, and gradient clipping at 5.0.
The principal configuration and split protocol are reported here to make the
main empirical claims interpretable without supplementary material.

\begin{table}[t]
\caption{Event counts induced by the 50-snapshot \method{} protocol.
The 70/15/15 split is performed by snapshot index, so event percentages need
not be exactly 70/15/15}
\label{tab:datasets}
\centering
\small
\setlength{\tabcolsep}{5pt}
\begin{tabular}{lrrrr}
\toprule
Dataset & Events & Train & Validation & Test \\
\midrule
MOOC      & 411,749 & 266,357 & 68,320  & 77,072 \\
Wikipedia & 157,474 & 110,947 & 25,099  & 21,428 \\
Reddit    & 672,447 & 468,326 & 100,623 & 103,498 \\
\bottomrule
\end{tabular}
\end{table}

\subsection{Predictive Performance}

Table~\ref{tab:dsf_results} reports the three validation-selected checkpoints.
Best epochs range from 60 to 96, indicating that convergence occurs late under
the adopted optimization schedule.

\begin{table}[t]
\caption{Validation-selected \method{} results from 100-epoch runs}
\label{tab:dsf_results}
\centering
\small
\setlength{\tabcolsep}{5pt}
\begin{tabular}{lrrrrr}
\toprule
Dataset & Best ep. & Val AP & Test AUC & Test AP & Params \\
\midrule
MOOC      & 96 & 0.8080 & 0.7898 & 0.7851 & 93,350 \\
Wikipedia & 60 & 0.9183 & 0.9056 & 0.9088 & 132,998 \\
Reddit    & 87 & 0.9864 & 0.9851 & 0.9860 & 132,486 \\
\bottomrule
\end{tabular}
\end{table}

\paragraph{Comparison with dynamic spectral filtering.}
We additionally adapt the official DEFT implementation
\cite{bastos2023deft} to the same chronological rolling-prefix and fixed-negative
snapshot protocol. Table~\ref{tab:deft_comparison} reports the adopted
validation-selected \method{} checkpoint and the mean and standard deviation of
three DEFT seeds. \method{} is stronger on MOOC, DEFT is stronger on
Wikipedia, and their Reddit AP differs by less than 0.001. Because the
\method{} entry is a single adopted seed, this table supports a performance
comparison but not a paired statistical significance claim.

\begin{table}[t]
\caption{Predictive comparison with the official DEFT implementation under the
shared snapshot adapter. DEFT values are mean $\pm$ standard deviation over
three seeds; \method{} values are the adopted validation-selected seed-0 runs}
\label{tab:deft_comparison}
\centering
\small
\setlength{\tabcolsep}{4pt}
\begin{tabular}{lrrrr}
\toprule
Dataset & \method{} AUC & \method{} AP & DEFT AUC & DEFT AP \\
\midrule
MOOC & 0.7898 & \textbf{0.7851} & $0.7997{\pm}0.0062$ & $0.7717{\pm}0.0183$ \\
Wikipedia & 0.9056 & 0.9088 & $0.9281{\pm}0.0109$ & $\mathbf{0.9237{\pm}0.0142}$ \\
Reddit & 0.9851 & 0.9860 & $0.9868{\pm}0.0015$ & $\mathbf{0.9869{\pm}0.0017}$ \\
\bottomrule
\end{tabular}
\end{table}

\subsection{Component Ablation}

Table~\ref{tab:ablation} reports the existing controlled MOOC component
ablation. All rows use the same short training budget, data pipeline,
Chebyshev propagation, decoder, and loss; the table is intended to isolate
architectural contributions and is not substituted for the converged result in
Table~\ref{tab:dsf_results}.

\begin{table}[t]
\caption{Controlled MOOC component ablation under a shared short training
budget. The converged main result is reported separately}
\label{tab:ablation}
\centering
\small
\setlength{\tabcolsep}{4pt}
\begin{tabular}{lrrr}
\toprule
Variant & AUC & AP & Params \\
\midrule
Time-conditioned DSF & 0.6565 & 0.6196 & 34,785 \\
$+$ mean history & 0.6631 & 0.6426 & 41,025 \\
$+$ filter GRU, no gate & 0.6878 & 0.6687 & 76,513 \\
$+$ single global gate & 0.6941 & 0.6735 & 84,834 \\
$+$ order-only gate & 0.6628 & 0.6330 & 85,029 \\
$+$ hierarchical global--order gate & \textbf{0.7105} & \textbf{0.6980} & 93,350 \\
\bottomrule
\end{tabular}
\end{table}

Replacing independently time-conditioned coefficients with recurrent
coefficient evolution improves AP by 0.0492, supporting the central claim that
the propagation response benefits from persistent temporal state. A scalar
gate adds a further 0.0048 AP. An order-specific gate alone performs worse,
whereas the hierarchical factorization improves AP by 0.0245 over the scalar
gate and by 0.0293 over the ungated GRU. This pattern supports the intended
division of labor: the global gate limits overall drift, while local gates
determine which spectral orders may change.

\subsection{Computational Efficiency}

Table~\ref{tab:resource} compares the measured resource footprint of
\method{}, TGN, TGAT, DyGFormer, and DEFT on the same workstation. Training time is
the mean wall-clock time per epoch and excludes validation and final-test
inference. The adopted \method{} timings come from the 100-epoch runs; TGAT and
DyGFormer were timed in three-epoch seed-0 runs, while TGN uses the corresponding
official-implementation benchmark run. DEFT uses the official core
implementation with a task adapter and is averaged over three formal seeds.
Because a \method{} epoch processes
snapshots whereas the other models process event minibatches, time per epoch is
an empirical end-to-end measurement rather than an architecture-independent
complexity unit.

\begin{table}[t]
\caption{Measured computational efficiency. Time is mean training seconds per
epoch; GPU is peak allocated memory in MB. Lower is better for all columns}
\label{tab:resource}
\centering
\small
\setlength{\tabcolsep}{5pt}
\begin{tabular}{llrrr}
\toprule
Dataset & Model & Time/epoch (s) & Params & GPU MB \\
\midrule
\multirow{5}{*}{MOOC}
 & \method{} & \textbf{1.766} & \textbf{93,350} & \textbf{130} \\
 & DEFT & 14.211 & 804,231 & 4,226 \\
 & TGN & 689.257 & 1,014,457 & 1,037 \\
 & TGAT & 370.741 & 2,578,625 & 2,824 \\
 & DyGFormer & 643.486 & 1,259,935 & 2,234 \\
\midrule
\multirow{5}{*}{Wikipedia}
 & \method{} & \textbf{1.607} & \textbf{132,998} & \textbf{68} \\
 & DEFT & 8.475 & 1,100,751 & 2,113 \\
 & TGN & 23.796 & 1,216,729 & 892 \\
 & TGAT & 141.423 & 2,578,625 & 2,628 \\
 & DyGFormer & 106.196 & 1,087,035 & 1,206 \\
\midrule
\multirow{5}{*}{Reddit}
 & \method{} & \textbf{2.103} & \textbf{132,486} & \textbf{182} \\
 & DEFT & 40.513 & 1,100,239 & 4,542 \\
 & TGN & 30.411 & 1,216,729 & 595 \\
 & TGAT & 618.658 & 2,578,625 & 3,150 \\
 & DyGFormer & 432.270 & 1,111,735 & 1,924 \\
\bottomrule
\end{tabular}
\end{table}

Across the three datasets, \method{} uses $8.2$--$27.6\times$ fewer parameters
and $3.3$--$38.6\times$ less peak GPU memory than the measured alternatives.
The advantage is consistent rather than dataset specific: \method{} is the
smallest and least memory-intensive model in every dataset block. Even on
Reddit, the largest dataset considered, it trains with 132K parameters and
182 MB, compared with 1.11M parameters and 1.92 GB for DyGFormer and 2.58M
parameters and 3.15 GB for TGAT.

The closest mechanism-level comparison is DEFT. Relative to its official
implementation, \method{} uses $8.3$--$8.6\times$ fewer parameters,
$25.0$--$32.5\times$ less allocated GPU memory, and $5.3$--$19.3\times$ less
time per epoch. Combined with Table~\ref{tab:deft_comparison}, this is the
central empirical result: directly evolving a compact coefficient state
preserves much of the predictive capability of a composite dynamic spectral
model while occupying a substantially better resource regime.

The wall-clock measurements show the same pattern. Relative to \method{}, the
observed time per epoch is $14.5$--$390.3\times$ higher for TGN,
$88.0$--$294.2\times$ higher for TGAT, and $66.1$--$364.4\times$ higher for
DyGFormer. This is consistent with the intended design: temporal adaptation is
concentrated in a graph-size-independent spectral state instead of node memory
or long-history attention. The result is not merely a smaller parameter file;
it reduces both runtime working memory and measured training cost. These
measurements establish a clear resource-footprint advantage under the
evaluated implementations. They do not by themselves imply a
controlled predictive-accuracy ranking because the snapshot and event-stream
pipelines use different split and negative-sampling granularities.

\subsection{Evaluation Protocol}

\method{} uses strictly chronological snapshot splits, rolling-prefix
evaluation, and same-source negative sampling that rejects current-bucket
positives. DEFT uses the same snapshot adapter and fixed negatives. The
event-stream baselines retain their native sampling granularity, so their
accuracy values should not be read as a perfectly controlled cross-family
ranking. These qualifications are applied consistently when interpreting the
reported cross-family comparisons.

\section{Limitations}

The current study has three principal limitations. First, \method{} uses
discrete temporal snapshots, while several baselines process individual
events; snapshot resolution is therefore an additional modeling choice.
Second, although DEFT is evaluated through the same chronological snapshot
adapter and fixed-negative protocol as \method{}, the event-stream baselines
do not share exactly the same split and negative-sampling implementation. A
definitive ranking across all model families requires a shared
timestamp-aware sampler. Third, the present evidence uses a single adopted seed
for the converged 100-epoch \method{} runs. Multi-seed converged experiments
would be needed for uncertainty estimates. Finally, an epoch has different
computational semantics for snapshot and event-stream implementations; the
reported wall-clock values are useful system measurements on common hardware,
but not normalized operation counts.

\section{Conclusion}

We studied an underexplored operator-centric perspective on temporal graph
learning and instantiated it through Dynamic Spectral Filtering. Rather than
claiming historical priority for dynamic spectral models, \method{} isolates a
focused mechanism: it turns the coefficients of a localized spectral
propagation rule into a recurrent, hierarchically gated temporal state. The
resulting model reaches AP values of
0.7851, 0.9088, and 0.9860 on three temporal link-prediction datasets with a
compact parameter and memory footprint. Relative to DEFT, the closest spectral
baseline, \method{} is better on MOOC, nearly tied on Reddit, and modestly
lower on Wikipedia while requiring dramatically fewer resources. The current
evidence therefore supports an accuracy--efficiency result: the evaluated
\method{} runs require 93K--133K
parameters, 68--182 MB peak GPU memory, and 1.6--2.1 seconds per epoch, all
substantially below the measured alternatives. This consistency across all
three datasets suggests that evolving a compact propagation response can offer
a practical alternative when node-indexed memory or long-history attention is
computationally prohibitive. Controlled
cross-family accuracy ranking requires a unified evaluation protocol.

\section*{Reproducibility Statement}
The causal snapshot construction, evaluation protocol, principal
hyperparameters, and model dimensions are described in
Section~\ref{sec:experiments}. The implementation used for the reported
results will be released with the paper.

{\footnotesize
\bibliography{references}
\bibliographystyle{unsrtnat}
}

\end{document}